\def\year{2022}\relax
\documentclass[letterpaper]{article} % DO NOT CHANGE THIS
\usepackage{aaai22}  % DO NOT CHANGE THIS
\usepackage{times}  % DO NOT CHANGE THIS
\usepackage{helvet}  % DO NOT CHANGE THIS
\usepackage{courier}  % DO NOT CHANGE THIS
\usepackage[hyphens]{url}  % DO NOT CHANGE THIS
\usepackage{graphicx} % DO NOT CHANGE THIS
\usepackage{natbib}  % DO NOT CHANGE THIS AND DO NOT ADD ANY OPTIONS TO IT
\usepackage{caption} % DO NOT CHANGE THIS AND DO NOT ADD ANY OPTIONS TO IT
\DeclareCaptionStyle{ruled}{labelfont=normalfont,labelsep=colon,strut=off} % DO NOT CHANGE THIS
\usepackage{algorithm}
\usepackage{algorithmic}
\usepackage{newfloat}
\usepackage{listings}
\floatstyle{ruled}
\newfloat{listing}{tb}{lst}{}
\floatname{listing}{Listing}

\usepackage{leftidx}
\usepackage{amsmath}
\usepackage{amssymb}
\usepackage{mathtools}
\usepackage{bbding}
\usepackage{booktabs}
\usepackage{multirow}
\usepackage{makecell}
\usepackage{graphicx}
\usepackage{enumitem}
\usepackage{wrapfig}
\usepackage{xcolor}

\usepackage[switch]{lineno}  % add line numbers

\DeclarePairedDelimiter\norm{\lVert}{\rVert_2}
\DeclarePairedDelimiter\abs{|}{|}
\newcommand{\tr}[1]{{\textcolor{red}{#1}}}
\newcommand{\tb}[1]{{\textcolor{blue}{#1}}}
\title{FINet: Dual Branches Feature Interaction for \\Partial-to-Partial Point Cloud Registration}
\author{
Hao Xu$^{1,2}$, 
Nianjin Ye$^{2}$,
Guanghui Liu$^1$\thanks{Corresponding author},
Bing Zeng$^1$,
Shuaicheng Liu$^{1,2*}$
}
\affiliations{
\textsuperscript{1}University of Electronic Science and Technology of China \\
{\tt \{xuhao02@std., guanghuiliu@, eezeng@, liushuaicheng@\}uestc.edu.cn}\\
\textsuperscript{2}Megvii Technology \\
{\tt \{xuhao02, yenianjin, liushuaicheng\}@megvii.com}
}
\usepackage{bibentry}
\begin{document}

\maketitle
% \linenumbers

\begin{abstract}
Data association is important in the point cloud registration. In this work, we propose to solve the partial-to-partial registration from a new perspective, by introducing multi-level feature interactions between the source and the reference clouds at the feature extraction stage, such that the registration can be realized without the attentions or explicit mask estimation for the overlapping detection as adopted previously. Specifically, we present FINet, a feature interaction-based structure with the capability to enable and strengthen the information associating between the inputs at multiple stages. To achieve this, we first split the features into two components, one for rotation and one for translation, based on the fact that they belong to different solution spaces, yielding a dual branches structure. Second, we insert several interaction modules at the feature extractor for the data association. Third, we propose a transformation sensitivity loss to obtain rotation-attentive and translation-attentive features. Experiments demonstrate that our method performs higher precision and robustness compared to the state-of-the-art traditional and learning-based methods. Code is available at \url{https://github.com/megvii-research/FINet}.
\end{abstract}

\section{Introduction}
Point cloud registration is a longstanding research problem in the areas of computer vision and computer graphics, including augmented reality~\cite{azuma1997survey, billinghurst2015survey}, object pose estimation~\cite{fang2018learning, wang2019densefusion} and 3D reconstruction~\cite{lin2018learning}. It aims to predict a rigid 3D transformation, aligning the source point cloud to the reference. Data association is critical for aligning two point clouds, especially for the practical partial-to-partial scenes where inputs are obscured by partiality and contaminated by noise. Algorithms for this task have been improved steadily, which can be divided into two categories: correspondence matching-based and global feature-based methods. 

\begin{figure}[t]
    \centering
    \includegraphics[width=\linewidth]{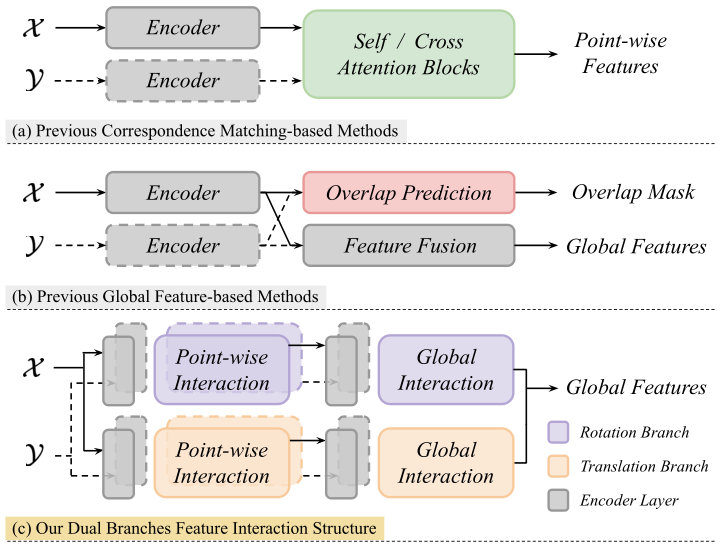}\\
    \caption{The structure comparison of partiality handling. The specific designs are shown in different colors. Our method use the multi-level feature interaction to deal with the partiality without using any attention mechanism in (a) or explicit overlapping region prediction in (b).}
    \label{fig:teaser}
\end{figure}
Iterative Closest Point (ICP)~\cite{besl1992method} is the most classical algorithm among the correspondence matching-based methods, where the correspondences are obtained by the nearest neighbor search. Subsequently, several methods~\cite{segal2009generalized, pomerleau2015review, yang2013go} are proposed to either improve the matching quality or search a larger transformation space. Recently, some learning-based approaches implement descriptor with neural network to improve the performance. Specifically, \cite{wang2019prnet} adopts Graph Neural Network (GNN)~\cite{wang2019dynamic} and attention mechanism~\cite{Vaswani2017}, which enables the information exchange between the inputs. However, most of them rely on the deep features that extracted from the local geometric structures. As a result, they can hardly utilize the knowledge of the entire point clouds. Without a global picture, the data association is inefficient.

In contrast, global feature-based methods can overcome the above-mentioned issues by aggregating global information without correspondences, e.g., PointNetLK~\cite{aoki2019pointnetlk} and Feature-metric Registration (FMR)~\cite{huang2020feature}. Although they can learn features from the entire point cloud, they perform poorly in partial-to-partial registration due to the lack of data association. Recently, OMNet~\cite{xu2021omnet} predicts overlapping masks, converting the partial-to-partial registration to the registration of the same shape. However, the mask is predicted based on the features that extracted without early information exchange, making it harder to be estimated accurately. 

% In this paper, we propose to solve the partial-to-partial registration from a new perspective, by introducing feature interactions between the input source and reference point clouds. We show that abundant information exchange between inputs at the feature extraction stage can naturally equip the network with the capability of the overlapping selection. As such, only overlapped points contribute to the final result. Interestingly, this can be achieved implicitly during the feature extraction, without the need of explicit mask estimation or attention modules, as long as we enable feature interactions.  

In this paper, we solve the partial-to-partial registration from a new perspective, by introducing multi-level feature interactions between the input point clouds. We show that abundant information exchange between the inputs at the feature extraction stage can naturally equip the network with the partiality perceptual capability. As such, the global features from two inputs can focus on the same parts of an object. Interestingly, as shown in Fig.~\ref{fig:teaser}, this can be achieved implicitly without the need of attention modules or explicit mask estimations, as long as feature interactions are enabled.

To this end, we propose FINet: a feature interaction-based structure with the ability to enable and strengthen the data association between the inputs at multiple stages. To promote the information associating, several interaction modules are inserted to the feature extractor. Besides, the 3D rigid transformation consists of 3D translation and 3D rotation, which reside in different solution spaces. Previously, they are regressed from the same deep feature. In this work, we implement a dual branches structure to process them separately so as to enhance the feature interactions. Moreover, based on the dual branches structure, we design a transformation sensitivity loss, which encourages the network to extract the rotation-attentive and translation-attentive features, improving the quality of regression from their own solution space. Furthermore, to avoid concentrating on local geometry, we propose a point-wise feature dropout loss to encourage aggregating features from scattered locations.
We summarize our key contributions as follows:
\begin{itemize}[itemsep=-1pt]
    \item We propose a multi-level feature interaction module for the point cloud registration, which promotes the data association between the inputs and shows state-of-the-art performance on several partially visible datasets.
    
    \item We propose a dual branches structure to mitigate the effect of solution space disparity between rotation and translation, so as to further enhance feature interactions. 
    
    \item We design a transformation sensitivity loss, which supervises two branches of the feature extractor to learn rotation-attentive and translation-attentive features. Besides, we propose a point-wise feature dropout loss to facilitate the learning of global information. 
\end{itemize}

\section{Related Work}
\noindent\textbf{Correspondence Matching-based Methods. }
ICP~\cite{besl1992method} and its variants~\cite{rusinkiewicz-normal-sampling, segal2009generalized} are the earlier correspondence matching-based methods, which calculate the nearest neighbors as correspondences. However, they are often strapped into local minima due to the non-convexity. To this end, Go-ICP~\cite{yang2013go} utilizes a branch-and-bound strategy to find a good optimum at the expense of speed. Symmetric ICP~\cite{rusinkiewicz2019symmetric} improves the point-to-plane objective function. Furthermore, Fast Global Registration (FGR)~\cite{zhou2016fast} uses FPFH~\cite{FPFH} features and an alternating optimization technique to improve efficiency. Recent learning-based methods use Multi-Layer Perceptron (MLP)~\cite{qi2017pointnet, qi2017pointnet++} or GNN~\cite{wang2019dynamic} to replace the handcrafted descriptors. Specifically, DCP~\cite{wang2019deep} calculates feature-to-feature correspondences. DeepGMR~\cite{yuan2020deepgmr} integrates Gaussian Mixture Model (GMM) to learn point-to-GMM correspondences. Generally, the accurate matching heavily relies on the distinctive geometric structures, and an extra time-consuming RANSAC~\cite{fischler1981random} may be needed. In contrast, we aggregate feature from the entire point clouds and achieve an end-to-end registration.
\begin{figure*}[t]
    \centering
    \includegraphics[width=\linewidth]{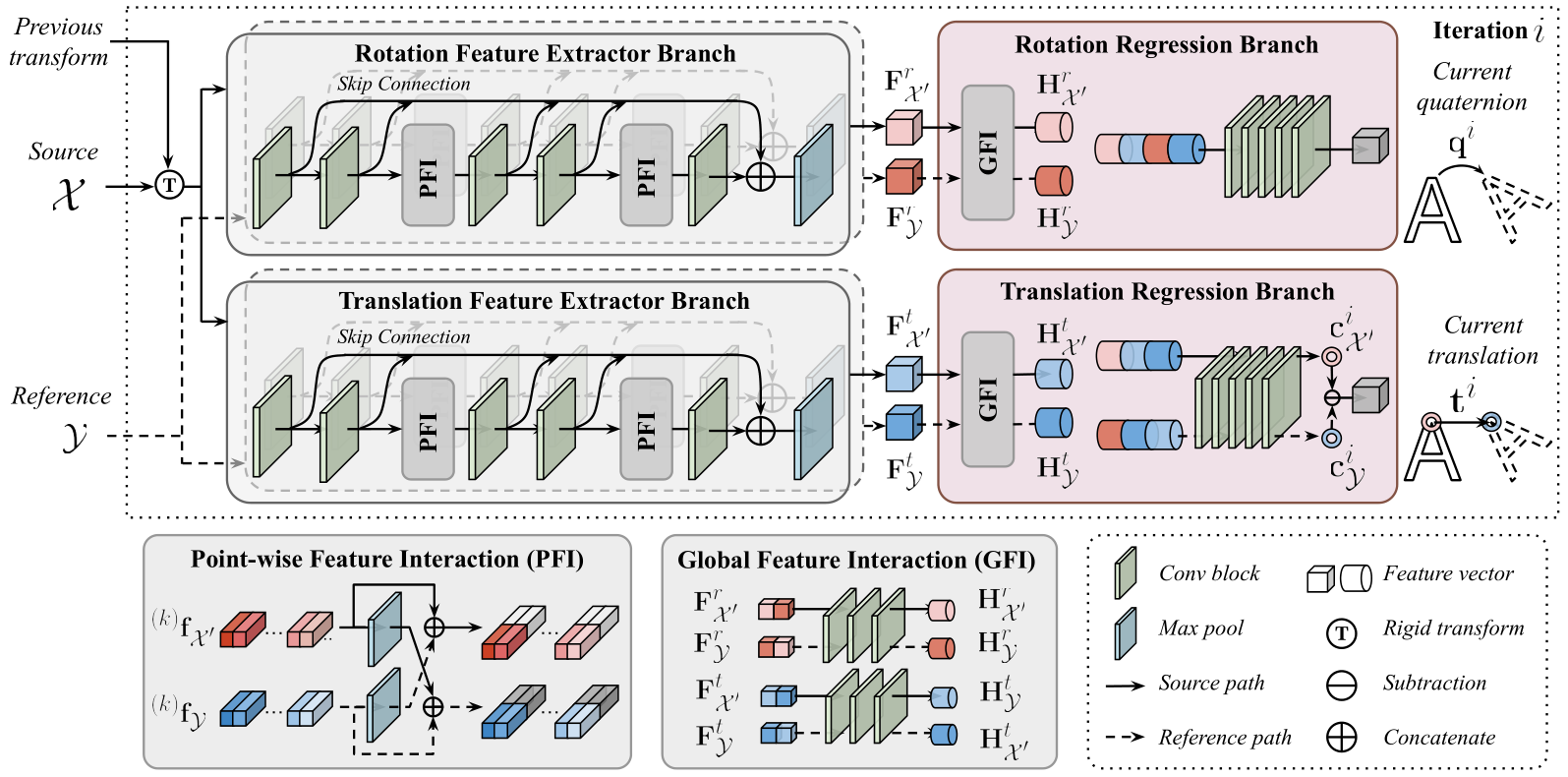}\\
    \caption{Network architecture for FINet and the multi-levels feature interaction modules.}
    \label{fig:pipeline}
\end{figure*}

\noindent\textbf{Global Feature-based Methods. }
PointNetLK~\cite{aoki2019pointnetlk} pioneers the global feature-based methods, which utilizes the Lucas \& Kanade (LK) algorithm~\cite{lucas1981iterative} to solve the rigid transformation. PCRNet~\cite{sarode2019pcrnet} improves the robustness to noise by replacing the LK algorithm with a MLP. Subsequently, FMR~\cite{huang2020feature} constrains the global feature distance of the inputs with an extra decoder. However, all of them ignore the effect of partiality. Our method is aware of the partiality and robust to different scenes.

\noindent\textbf{Partial-to-partial Registration. }
As a more realistic problem, partial-to-partial registration is studied by several works. Particularly, PRNet~\cite{wang2019prnet} uses the Gumble-Softmax~\cite{jang2016categorical} to improve the feature matching. RPMNet~\cite{yew2020-RPMNet} applies the Sinkhorn normalization~\cite{sinkhorn1964relationship} to encourage the bijectivity of the matching. RGM~\cite{fu2021robust} further utilizes deep graph matching to reject outliers. Meanwhile, some methods are designed for the more challenging indoor scenes, e.g., DGR~\cite{choy2020deep}, FCGF~\cite{choy2019fully} and D3Feat~\cite{bai2020d3feat}. Recently proposed PREDATOR~\cite{huang2020predator} predicts the overlapping and matchability scores to tackle the partiality. Similarly, OMNet~\cite{xu2021omnet} estimates the overlapping masks. Nevertheless, most of these methods lack early information exchange between the inputs, resulting in features that are imperceptible of partiality.% partiality-imperceptible features. % Besides, the accuracy of matching using deep feature descriptors heavily relies on the high point cloud density. In contrast, we designs simple yet efficient structures to enable the data association and registers with fewer input points.

% \vspace{-0.3cm}
\noindent\textbf{Feature Interaction. }
Some previous works implement the data association during the feature extraction. Particularly, DCP and PRNet introduce the attention module~\cite{Vaswani2017} to enable information exchanging between the inputs. Nevertheless, the 3D local features lack the global information, further reducing the effectiveness of the feature interaction. Although PREDATOR applies cross-attention mechanism to capture the global picture, it has significant computational and memory requirements. Besides, the data associations in OMNet is located after the feature extractor, which is too late to catch enough partiality knowledge. However, our method possesses feature interactions at multiple levels, which enables the partiality perception without any attention mechanism or overlapping prediction.
% \vspace{-1mm}

\section{Method} 
% \vspace{-1mm}
Fig.~\ref{fig:pipeline} shows an illustration of our method. We estimate the 3D rigid transformation iteratively, which is represented in the form of quaternion $\mathbf{q}$ and translation $\mathbf{t}$. The entire structure can be divided into two parts: a dual branches encoder (\S\ref{sec3.2}) and a dual branches transformation regressor (\S\ref{sec3.3}), where the multi-level features of the source and the reference point clouds are interactive with each other (\S\ref{sec3.4}). Finally, the loss functions are detailed explained (\S\ref{sec3.5}).

\subsection{Notation}
\label{sec3.1}
Here we introduce some notation that will be used throughout the paper. The registration problem is that for a given source point cloud $\mathcal{X} \in \mathbb{R}^{3 \times N_{\mathcal{X}}}$ and a reference point cloud $\mathcal{Y} \in \mathbb{R}^{3 \times N_{\mathcal{Y}}}$ of $N_{\mathcal{X}}$ and $N_{\mathcal{Y}}$ points, we aim to find the 3D rigid transformation $\{\mathbf{R},\mathbf{t}\}$ that aligns $\mathcal{X}$ to $\mathcal{Y}$.

\subsection{Dual Branches Encoder}
\label{sec3.2}
Since the translation belongs to Euclidean space,  which is little correlated to the quaternion space. It is not appropriate to obtain the rotation-attentive and translation-attentive features with shared weights. Meanwhile, the features that only contain the rotation or translation information are more beneficial to the data association. To this end, we use a dual branches encoder to extract features for them separately. Refer to~\cite{huang2020pf}, point-wise features that output from each convolution block in the MLP are extracted to combine the multi-level features by max-pooling. To promote the low-level information propagation between the source and the reference point clouds, two point-wise feature interaction modules are injected into the encoder. At each iteration, the source point cloud $\mathcal{X}$ is first transformed by the previous transformation into the transformed point cloud $\mathcal{X}^{\prime}$. The encoder takes $\mathcal{X}^{\prime}$ and the reference point cloud $\mathcal{Y}$ as inputs, generating the global features as follow:
\begin{equation}
    \small
    % rotation branch
    \mathbf{F}^{n}_{m}
    =\operatorname{max}(\operatorname{cat}[
    \{\leftidx{^{(k)}}{\mathbf{f}^{n}_{m}}|k=1..K\}
    ]).
\end{equation}
Here, $m\in\{\mathcal{X}^{\prime},\mathcal{Y}\}, \; n \in \{r,t\}$. The superscript $r$ and $t$ denote the global features $\mathbf{F}$ that belong to the rotation and translation encoders respectively. The superscript $k$ represents the point-wise features $\mathbf{f}$ that output from the $k$-th convolution block, and there are $K$ blocks in total. $\operatorname{max}(\cdot)$ denotes channel-wise max-pooling, and $\operatorname{cat}[\cdot,\cdot]$ means concatenation. The encoder is shared weights for the inputs.
\begin{table*}[t]
    \normalsize
    \resizebox{\linewidth}{!}{%
    \begin{tabular}{clcccccccccccc}
        \toprule
        & \multicolumn{1}{c}{} & \multicolumn{2}{c}{RMSE(\textbf{R})} & \multicolumn{2}{c}{MAE(\textbf{R})} & \multicolumn{2}{c}{RMSE(\textbf{t})} & \multicolumn{2}{c}{MAE(\textbf{t})} & \multicolumn{2}{c}{Error(\textbf{R})} & \multicolumn{2}{c}{Error(\textbf{t})} 
        \\ 
        \cmidrule(lr){3-4} \cmidrule(lr){5-6} \cmidrule(lr){7-8} \cmidrule(lr){9-10} \cmidrule(lr){11-12} \cmidrule(lr){13-14} 
        &\multicolumn{1}{c}{\multirow{-2}{*}{Method}}&\emph{OS} &\emph{TS} &\emph{OS} &\emph{TS} &\emph{OS} &\emph{TS} &\emph{OS} &\emph{TS} &\emph{OS} &\emph{TS} &\emph{OS} &\emph{TS}
        \\ 
        \midrule
        & ICP~\cite{besl1992method}            & 20.036     & 22.840     & 10.912     & 12.147     & 0.1893      & 0.1931      & 0.1191      & 0.1217      & 22.232     & 24.654     & 0.2597      & 0.2612      \\
        % & Go-ICP~\cite{yang2013go}             & 70.776     & 71.077     & 39.000     & 38.266     & 0.3111      & 0.3446      & 0.1807      & 0.1936      & 71.597     & 76.492     & 0.3996      & 0.4324      \\
        & Symmetric ICP~\cite{rusinkiewicz2019symmetric} & 10.419 & 11.295 & 8.992 & 9.592 & 0.1367 & 0.1394 & 0.1082 & 0.1124 & 17.954 & 19.571 & 0.2367 & 0.2414 \\
        & FGR~\cite{zhou2016fast}              & 48.533     & 46.766     & 29.661     & 29.635     & 0.2920      & 0.3041      & 0.1965      & 0.2078      & 55.855     & 57.685     & 0.4068      & 0.4263      \\
        & PointNetLK~\cite{aoki2019pointnetlk} & 23.866     & 27.482     & 15.070     & 18.627     & 0.2368      & 0.2532      & 0.1623      & 0.1778      & 29.374     & 36.947     & 0.3454      & 0.3691      \\
        & DCP~\cite{wang2019deep}              & 12.217     & 11.109     & 9.054      & 8.454      & 0.0695      & 0.0851      & 0.0524      & 0.0599      & 7.835      & 9.216      & 0.1049      & 0.1259      \\
        & RPMNet~\cite{yew2020-RPMNet}         & 1.347      & 2.162      & 0.759      & 1.135      & 0.0228      & 0.0267      & 0.0089      & 0.0141      & 1.446      & 2.280      & 0.0193      & 0.0302      \\
        & FMR~\cite{huang2020feature}          & 7.642      & 8.033      & 4.823      & 4.999      & 0.1208      & 0.1187      & 0.0723      & 0.0726      & 9.210      & 9.741      & 0.1634      & 0.1617      \\
        % & DeepGMR~\cite{yuan2020deepgmr}       & 72.310     & 70.886     & 49.769     & 47.853     & 0.3443      & 0.3703      & 0.2462      & 0.2582      & 82.652     & 86.444     & 0.5044      & 0.5354      \\
        & RGM~\cite{fu2021robust} & 3.470 & 4.912 & 1.251 & 1.786 & 0.0391 & 0.0428 & 0.0135 & 0.0183 & 2.441 & 3.506 & 0.0289 & 0.0393 \\
        & OMNet~\cite{xu2021omnet}             & \tb{0.771} & \tb{1.384} & \tb{0.277} & \tb{0.542} & \tb{0.0154} & \tb{0.0226} & \tb{0.0056} & \tb{0.0093} & \tb{0.561} & \tb{1.118} & \tb{0.0122} & \tb{0.0198} \\
        \multirow{-10}{*}{\rotatebox{90}{(a) Unseen Shapes}} 
        & Ours                                 & \tr{0.694} & \tr{1.267} & \tr{0.198} & \tr{0.269} & \tr{0.0076} & \tr{0.0168} & \tr{0.0029} & \tr{0.0048} & \tr{0.383} & \tr{0.591} & \tr{0.0066} & \tr{0.0110} \\
        \midrule
        & ICP~\cite{besl1992method}            & 20.387     & 22.906     & 12.651     & 13.599     & 0.1887      & 0.1994      & 0.1241      & 0.1286      & 25.085     & 26.819     & 0.2626      & 0.2700      \\
        % & Go-ICP~\cite{yang2013go}             & 69.747     & 64.455     & 39.646     & 34.017     & 0.3035      & 0.3196      & 0.1788      & 0.1888      & 68.329     & 68.920     & 0.3893      & 0.4091      \\
        & Symmetric ICP~\cite{rusinkiewicz2019symmetric} & 12.291 & 12.333 & 10.841 & 10.746 & 0.1488 & 0.1456 & 0.1212 & 0.1186 & 21.399 & 21.437 & 0.2577 & 0.2521 \\
        & FGR~\cite{zhou2016fast}              & 46.161     & 41.644     & 27.475     & 26.193     & 0.2763      & 0.2872      & 0.1818      & 0.1951      & 49.749     & 51.463     & 0.3745      & 0.4003      \\
        & PointNetLK~\cite{aoki2019pointnetlk} & 27.903     & 42.777     & 18.661     & 28.969     & 0.2525      & 0.3210      & 0.1752      & 0.2258      & 36.741     & 53.307     & 0.3671      & 0.4613      \\
        & DCP~\cite{wang2019deep}              & 13.387     & 12.507     & 9.971      & 9.414      & 0.0762      & 0.1020      & 0.0570      & 0.0730      & 11.128     & 12.102     & 0.1143      & 0.1493      \\
        & RPMNet~\cite{yew2020-RPMNet}         & 3.934      & 7.491      & 1.385      & 2.403      & 0.0441      & 0.0575      & \tb{0.0150} & 0.0258      & \tb{2.606} & 4.635      & \tb{0.0318} & 0.0556      \\
        & FMR~\cite{huang2020feature}          & 10.365     & 11.548     & 6.465      & 7.109      & 0.1301      & 0.1330      & 0.0816      & 0.0837      & 12.159     & 13.827     & 0.1773      & 0.1817      \\
        % & DeepGMR~\cite{yuan2020deepgmr}       & 75.773     & 68.425     & 53.689     & 46.269     & 0.3485      & 0.3667      & 0.2481      & 0.2595      & 85.210     & 87.192     & 0.5074      & 0.5323      \\
        & RGM~\cite{fu2021robust} &4.983 & 7.298 & 1.669 & 2.259 & 0.0402 & 0.0624 & 0.0164 & 0.0234 & 3.254 & 4.474 & 0.0348 & 0.0511 \\
        & OMNet~\cite{xu2021omnet}             & \tb{3.719} & \tb{4.014} & \tb{1.314} & \tb{1.619} & \tb{0.0392} & \tb{0.0406} & 0.0151      & \tb{0.0179} & 2.657      & \tb{3.206} & 0.0321      & \tb{0.0383} \\
        \multirow{-10}{*}{\rotatebox{90}{(b) Unseen Categories}}
        & Ours                                 & \tr{3.583} & \tr{3.918} & \tr{1.109} & \tr{1.286} & \tr{0.0324} & \tr{0.0404} & \tr{0.0115} & \tr{0.0142} & \tr{2.207} & \tr{2.572} & \tr{0.0245} & \tr{0.0311} \\
        \midrule
        & ICP~\cite{besl1992method}            & 20.566     & 21.893     & 12.786     & 13.402     & 0.1917      & 0.1963      & 0.1265      & 0.1278      & 25.417     & 26.632     & 0.2667      & 0.2679      \\
        % & Go-ICP~\cite{yang2013go}             & 70.417     & 65.402     & 40.303     & 34.988     & 0.3072      & 0.3233      & 0.1822      & 0.1929      & 69.175     & 71.054     & 0.3962      & 0.4170      \\
        & Symmetric ICP~\cite{rusinkiewicz2019symmetric} & 12.183 & 12.576 & 10.723 & 10.987 & 0.1487 & 0.1478 & 0.1210 & 0.1203 & 21.169 & 21.807 & 0.2576 & 0.2560 \\
        & FGR~\cite{zhou2016fast}              & 49.133     & 46.213     & 31.347     & 30.116     & 0.3002      & 0.3034      & 0.2068      & 0.2141      & 56.652     & 58.968     & 0.4230      & 0.4364      \\
        & PointNetLK~\cite{aoki2019pointnetlk} & 26.476     & 29.733     & 19.258     & 21.154     & 0.2542      & 0.2670      & 0.1853      & 0.1937      & 37.688     & 42.027     & 0.3831      & 0.3964      \\
        & DCP~\cite{wang2019deep}              & 13.117     & 12.730     & 9.741      & 9.556      & 0.0779      & 0.1072      & 0.0591      & 0.0774      & 11.350     & 12.173     & 0.1187      & 0.1586      \\
        & RPMNet~\cite{yew2020-RPMNet}         & 4.118      & 6.160      & 1.589      & 2.467      & 0.0467      & 0.0618      & 0.0175      & 0.0274      & \tb{2.983} & 4.913      & 0.0378      & 0.0589      \\
        & FMR~\cite{huang2020feature}          & 10.604     & 11.674     & 6.725      & 7.400      & 0.1300      & 0.1364      & 0.0827      & 0.0867      & 12.627     & 14.121     & 0.1788      & 0.1870      \\
        % & DeepGMR~\cite{yuan2020deepgmr}       & 75.257     & 68.560     & 53.470     & 46.579     & 0.3509      & 0.3735      & 0.2519      & 0.2654      & 84.121     & 87.104     & 0.5180      & 0.5455      \\
        & RGM~\cite{fu2021robust} & 5.968 & 6.893 & 2.479 & 3.068 & 0.0583 & 0.0650 & 0.0247 & 0.0311 & 4.766 & 6.243 & 0.0515 & 0.0662 \\
        & OMNet~\cite{xu2021omnet}             & \tr{3.572} & \tb{4.356} & \tb{1.570} & \tb{1.924} & \tb{0.0391} & \tb{0.0486} & \tb{0.0172} & \tb{0.0223} & 3.073      & \tb{3.834} & \tb{0.0359} & \tb{0.0476} \\
        \multirow{-10}{*}{\rotatebox{90}{\makecell[c]{(c) Gaussian Noise}}}
        & Ours                                 & \tb{3.676} & \tr{3.841} & \tr{1.363} & \tr{1.532} & \tr{0.0327} & \tr{0.0379} & \tr{0.0130} & \tr{0.0158} & \tr{2.673} & \tr{2.984} & \tr{0.0273} & \tr{0.0336} \\
        \bottomrule
    \end{tabular}%
    }
    \caption{Results on ModelNet40 (using the partial manner of RPMNet). For each metric, \emph{OS} and \emph{TS} denote the results on the once-sampled and twice-sampled data. Red and blue indicate the best and the second-best performance.}
    \label{tab:main}
\end{table*}

\subsection{Dual Branches Regression}
\label{sec3.3}
Given the distinctive features $\mathbf{F}^{r}$ and $\mathbf{F}^{t}$ of the inputs, we use a global feature interaction module to produce the hybrid features $\mathbf{H}^{r}$ and $\mathbf{H}^{t}$ for the rotation and translation respectively. Consistent with the encoder, a dual branches regression network is applied to regress the parameters for the rotation and the translation separately. Specifically, the rotation regression branch takes all the global features as inputs and produces a 4D vector, which represents the 3D rotation $\mathbf{R}$ in the form of quaternion~\cite{shoemake1985animating} $\mathbf{q} \in \mathbb{R}^{4}, \mathbf{q}^{T} \mathbf{q}=1$. Meanwhile, rather than regressing the translation vector $\mathbf{t} \in \mathbb{R}^{3}$ directly, the translation regression branch produces two 3D vectors, which represent two saliency points of the source and the reference point clouds respectively, then calculating the difference between them as $\mathbf{t}$. At each iteration, the transformation $\{\mathbf{q},\mathbf{t}\}$ is obtained as
\begin{equation}
\begin{gathered}
    \small
    % rotation q
    \mathbf{q}=f^{r}_{\theta}(\operatorname{cat}[
    \mathbf{H}^{r}_{\scriptstyle \mathcal{X}^{\prime}},
    \mathbf{H}^{t}_{\scriptstyle \mathcal{X}^{\prime}},
    \mathbf{H}^{r}_{\scriptstyle \mathcal{Y}},
    \mathbf{H}^{t}_{\scriptstyle \mathcal{Y}}])
    % white space
    \text{,}\quad
    % translation t
    \mathbf{t}=\mathbf{c}_{\scriptstyle \mathcal{Y}}-\mathbf{c}_{\scriptstyle \mathcal{X}^{\prime}}.
    % src center
    \\
    \mathbf{c}_{\scriptstyle \mathcal{X}^{\prime}}=
    f^{t}_{\theta}(\operatorname{cat}[
    \mathbf{H}^{r}_{\scriptstyle \mathcal{X}^{\prime}},
    \mathbf{H}^{t}_{\scriptstyle \mathcal{X}^{\prime}},
    \mathbf{H}^{t}_{\scriptstyle \mathcal{Y}}]),
    % ref center
    \\
    \mathbf{c}_{\scriptstyle \mathcal{Y}}=
    f^{t}_{\theta}(\operatorname{cat}[
    \mathbf{H}^{r}_{\scriptstyle \mathcal{Y}},
    \mathbf{H}^{t}_{\scriptstyle \mathcal{Y}},
    \mathbf{H}^{t}_{\scriptstyle \mathcal{X}^{\prime}}]).
\end{gathered}
\end{equation}
Here, the functions $f^{r}_{\theta}(\cdot)$ and $f^{t}_{\theta}(\cdot)$ denote the rotation and translation regression networks. The vectors $\mathbf{c}_{\scriptstyle \mathcal{X}^{\prime}} \in \mathbb{R}^{3}$ and $\mathbf{c}_{\scriptstyle \mathcal{Y}} \in \mathbb{R}^{3}$ mean the coordinates of the saliency points.

\subsection{Multi-level Feature Interaction}
\label{sec3.4}
For partial-to-partial registration, the network has to possess partiality perception ability, which means the feature extraction of the inputs need the information from each other.

% \vspace{-0.3cm}
\noindent\textbf{Point-wise Feature Interaction. }
In the dual branches encoder, we insert the Point-wise Feature Interaction (PFI) after multiple convolution blocks. The point-wise features of one point cloud are first aggregated by channel-wise max-pooling, then broadcasted and concatenated with the point-wise features of the other point cloud at the same level. The encoder features are then updated as
\begin{equation}
    \small
    \leftidx{^{(k)}}{\mathbf{f}_{\scriptstyle \mathcal{X}^{\prime}}}
    =\leftidx{^{(k)}}{g_{\theta}}(\operatorname{cat}[
    \leftidx{^{(k-1)}}{\mathbf{f}_{\scriptstyle \mathcal{X}^{\prime}}},\,
    \operatorname{repeat}({\operatorname{max}(\leftidx{^{(k-1)}}{\mathbf{f}_{\scriptstyle \mathcal{Y}}}), N_{\mathcal{X}^{\prime}})}
    ]),
\end{equation}
where the function $\leftidx{^{(k)}}{g_{\theta}}(\cdot)$ denotes the $k$-th convolution block, and $\operatorname{repeat}(\mathbf{z},N)$ denotes repeating $N$ times for the vector $\mathbf{z}$ at the element-wise dimension. The positions of inputs are inverted when extracting features for $\mathcal{Y}$.

% \vspace{-0.3cm}
\noindent\textbf{Global Feature Interaction. }
Before regressing the transformation parameters, we further strengthen the information exchanging between the inputs by the Global Feature Interaction (GFI). The rotation and translation features are concatenated separately and sent into MLPs to generate the hybrid global features. The entire process is defined as
\begin{equation}
    \small
    % rotation
    \mathbf{H}^{r}_{\scriptstyle \mathcal{X}^{\prime}}
    =h^{r}_{\theta}(\operatorname{cat[
    \mathbf{F}^{r}_{\scriptstyle \mathcal{X}^{\prime}}, 
    \mathbf{F}^{r}_{\scriptstyle \mathcal{Y}}
    ]})
    % white space
    ,\quad
    % translation
    \mathbf{H}^{t}_{\scriptstyle \mathcal{X}^{\prime}}
    =h^{t}_{\theta}(\operatorname{cat[
    \mathbf{F}^{t}_{\scriptstyle \mathcal{X}^{\prime}}, 
    \mathbf{F}^{t}_{\scriptstyle \mathcal{Y}}
    ]}),
\end{equation}
where $h^{r}_{\theta}(\cdot)$ and $h^{t}_{\theta}(\cdot)$ denote the GFI functions for the rotation and translation . The positions of inputs are inverted when producing features for $\mathcal{Y}$.

\subsection{Loss Functions}
\label{sec3.5}
\noindent\textbf{Transformation Sensitivity Loss. }
To modify the sensitivity of the encoder to the 3D transformation, we design the transformation sensitivity loss, which is a variant of the triplet loss~\cite{dong2018triplet}. It follows a simple intuition: the rotation branch should be more attentive to rotation and less attentive to translation, and vice versa for the translation branch. Hence, we cast the features of transformed source point cloud $\mathbf{F}^{r}_{\scriptstyle \mathcal{X}^{\prime}}$ and $\mathbf{F}^{t}_{\scriptstyle \mathcal{X}^{\prime}}$ as anchor features. For the rotation, $\mathcal{X}^{\prime}$ is first rotated by the previous predicted $\mathbf{q}$ to form $\mathcal{X}^{\prime}_{r}$, then sent into the dual branches encoder to extract features $\mathbf{F}^{r}_{\scriptstyle \mathcal{X}^{\prime}_{r}}$ and $\mathbf{F}^{t}_{\scriptstyle \mathcal{X}^{\prime}_{r}}$. Similarly, using $\mathbf{t}$ to alternate $\mathbf{q}$, we can obtain the translated point cloud $\mathcal{X}^{\prime}_{t}$ and its features $\mathbf{F}^{r}_{\scriptstyle \mathcal{X}^{\prime}_{t}}$ and $\mathbf{F}^{t}_{\scriptstyle \mathcal{X}^{\prime}_{t}}$. The loss function is $\mathcal{L}_{s}=\mathcal{L}^{r}_{s}+\mathcal{L}^{t}_{s}$, where
\begin{equation}
    \small
    \mathcal{L}^{r}_{s}=\operatorname{max}(\,
    \norm{\mathbf{F}^{r}_{\scriptstyle \mathcal{X}^{\prime}}-\mathbf{F}^{r}_{\scriptstyle \mathcal{X}^{\prime}_{t}}}-
    \norm{\mathbf{F}^{r}_{\scriptstyle \mathcal{X}^{\prime}}-\mathbf{F}^{r}_{\scriptstyle \mathcal{X}^{\prime}_{r}}}+
    \delta,
    \;
    \norm{\mathbf{F}^{r}_{\scriptstyle \mathcal{X}^{\prime}}-\mathbf{F}^{r}_{\scriptstyle \mathcal{X}^{\prime}_{t}}}
    ).
\end{equation}
Here, $\delta=0.01$ denotes the margin. $\mathcal{L}^{t}_{s}$ is calculated with the features that output from the translation branch and exchanging the positive and negative pairs. 
% Note that it is harder for the encoder to produce distinctive features with persistent iteration, which is because that the pose difference between $\mathcal{X}^{\prime}$ and $\mathcal{Y}$ gradually becomes small. Therefore, we reduce the margin $\delta$ after the first iteration. 

% \vspace{-0.3cm}
\noindent\textbf{Point-wise Feature Dropout Loss. }
To avoid the case that the feature extractor only concentrates on the local geometry of the inputs, we design a point-wise feature dropout loss, which encourages the network to learn the global features from more dispersed regions of the inputs. Concretely, inspired by \cite{baldi2013understanding}, we randomly set some of the point-wise features to zero before the max-pool layers in the encoder, and the distances between the global features that calculated before and after the dropout operation are constrained by the loss $\mathcal{L}_{d}=\mathcal{L}^{\mathcal{X}^{\prime}}_{d}+\mathcal{L}^{\mathcal{Y}}_{d}$, where
\begin{equation}
    \small
    \mathcal{L}^{\mathcal{X}^{\prime}}_{d}=
    \norm{\mathbf{F}^{r}_{\scriptstyle \mathcal{X}^{\prime}}-\mathbf{F}^{r}_{\scriptstyle \mathcal{X}^{\prime}_{d}}}+
    \norm{\mathbf{F}^{t}_{\scriptstyle \mathcal{X}^{\prime}}-\mathbf{F}^{t}_{\scriptstyle \mathcal{X}^{\prime}_{d}}}.
    % +\norm{\mathbf{F}^{r}_{\scriptstyle \mathcal{Y}}-\mathbf{F}^{r}_{\scriptstyle \mathcal{Y}_{d}}}
    % +\norm{\mathbf{F}^{t}_{\scriptstyle \mathcal{Y}}-\mathbf{F}^{t}_{\scriptstyle \mathcal{Y}_{d}}},
\end{equation}
Here, the subscript $d$ denotes the features are processed by dropout, and the superscript $\mathcal{X}^{\prime}$ and $\mathcal{Y}$ denote the features belong to $\mathcal{X}^{\prime}$ and $\mathcal{Y}$. The dropout ratio is set to 0.3.

% \vspace{-0.3cm}
\noindent\textbf{Parameter Regression Loss. }
Following \cite{xu2021omnet}, we directly measure the deviation of $\{\mathbf{q}, \mathbf{t}\}$ from the ground truth. The 3D transformation parameter regression loss is
\begin{equation}
    \small
    \mathcal{L}_{p}=
    \abs{\mathbf{q}-\mathbf{q}_{gt}}+
    \lambda \,\norm{\mathbf{t}-\mathbf{t}_{gt}},
\end{equation}
where subscript $gt$ denotes the ground-truth. We notice that using the combination of $\ell^{1}$ and $\ell^{2}$ distance can marginally improve the performance. Besides, the factor $\lambda$ is empirically set to 4.0 in all our experiments.

Combining the terms above after $N$ iterations, we have the weighted sum loss
\begin{equation}
    \small
    \mathcal{L}_{total}=\frac{1}{N}\sum_{i=1}^{N}{(
    \mathcal{L}^{i}_{p}+
    \beta \, \mathcal{L}^{i}_{s}+
    \gamma \, \mathcal{L}^{i}_{d}
    )},
    % \qquad \text{where} \qquad
    % \beta,\gamma \in \mathbb{R}.
\end{equation}
where the factors $\beta$ and $\gamma$ are set to $10^{-3}$.

\section{Experiments} 
\subsection{Dataset and Implementation Details}
\label{sec4.1}

\begin{table}[t]
    \normalsize
    \resizebox{\linewidth}{!}{%
    \begin{tabular}{lcccccc}
        \toprule
        Method & RMSE(\textbf{R}) & MAE(\textbf{R}) & RMSE(\textbf{t}) & MAE(\textbf{t}) & Error(\textbf{R}) & Error(\textbf{t}) 
        \\ 
        \midrule
        ICP                      & 18.588     & 9.628      & 0.0920     & 0.0521     & 18.720      & 0.1026 \\
        Go-ICP                   & 15.214     & 4.650      & 0.0566     & 0.0223     & 9.002       & 0.0445 \\
        Symmetric ICP & 7.096    & 6.280      & 0.0688     & 0.0617     & 12.531      & 0.1191 \\
        FGR                      & 33.723     & 19.268     & 0.1593     & 0.0914     & 35.971      & 0.1828 \\
        PointNetLK               & 29.747     & 18.550     & 0.1841     & 0.1081     & 32.760      & 0.1959 \\
        DCP                      & 7.300      & 4.378      & 0.0389     & 0.0272     & 8.853       & 0.0539 \\
        PRNet                    & 5.883      & 3.037      & 0.0380     & 0.0237     & 5.974       & 0.0472 \\
        FMR                      & 5.304      & 2.779      & 0.0323     & 0.0172     & 5.392       & 0.0342 \\
        DeepGMR                  & 24.908     & 13.611     & 0.1057     & 0.0689     & 25.830      & 0.1371 \\
        IDAM                     & 8.008      & 4.559      & 0.0484     & 0.0291     & 8.774       & 0.0578 \\
        OMNet                    & \tb{2.563} & \tb{1.215} & \tb{0.0183}& \tb{0.0098}& \tb{2.360}  & \tb{0.0196} \\
        Ours                     & \tr{2.378} & \tr{1.039} & \tr{0.0152}& \tr{0.0068}& \tr{2.018}  & \tr{0.0138} \\
        \bottomrule
    \end{tabular}%
    }
    \caption{Results on the TS unseen categories with Gaussian noise (using the partial manner of PRNet).}
    \label{tab:different_partial_manner}
\end{table}

\noindent\textbf{ModelNet40. }
\cite{wu20153d} includes CAD models from 40 man-made object categories. We use the processed data from OMNet~\cite{xu2021omnet}, where 8 axisymmetrical categories are removed to avoid the ill-posed problem. We denote the data that sampled once from the CAD model as once-sampled (\emph{OS}), while sampled twice separately as twice-sampled (\emph{TS}). We use the official train/test splits, resulting in 4,196 training, 1,002 validation, and 1,146 test objects. Two manners proposed by PRNet and RPMNet are applied to generate partially visible data. 
% Different from OMNet, we mainly show results on the more challenging RPMNet data.

\noindent\textbf{7Scenes. }
\cite{shotton2013scene} is a widely used benchmark where data is captured by a Kinect camera in 7 indoor scenes. We use the code from~\cite{zeng20173dmatch} to generate scan pairs with $>$30\% overlap, and adopt the official train/test splits, resulting in 8,042 training and 3,389 test pairs. We random sample 2,048 points from each scan to train our model, and set the voxel sample grid to 0.03 for training other deep learning-based methods.
\begin{figure}[h]
    \centering
    \includegraphics[width=\linewidth]{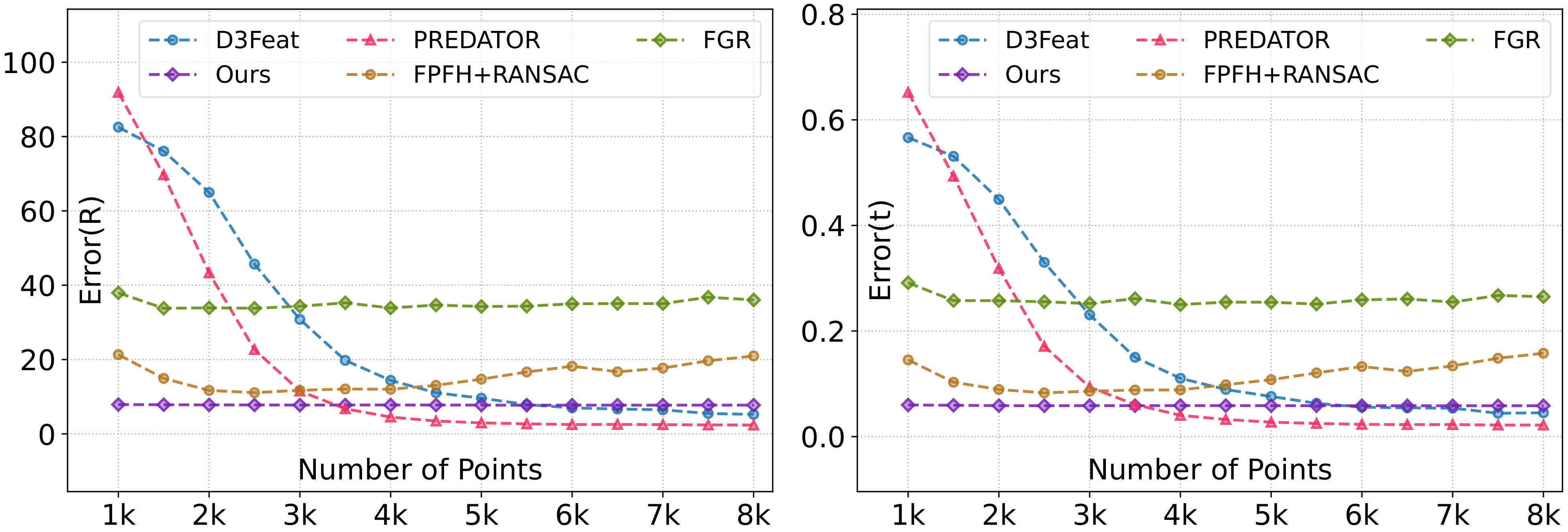}\\
    \caption{Errors on 7Scenes with different number of points.}
    \label{fig:7scenes_diff_points}
\end{figure}
\noindent\textbf{Implementation Details. }
We run 4 iterations of alignment. Adam optimizer~\cite{kingma2014adam} is used with $l_r=10^{-4}$. The batch size is 64, and training for 260k steps. 

\subsection{Baseline Algorithms}
\label{sec4.2}
We compare our method to the traditional methods: ICP (point-to-point), Go-ICP, Symmetric ICP and FGR, and the learning-based works: PointNetLK, DCP, RPMNet, FMR (supervised version), PRNet, IDAM, DeepGMR, RGM, OMNet, D3Feat and PREDATOR. We use the implementations of ICP and FGR in Intel Open3D~\cite{zhou2018open3d}, Symmetric ICP in PCL~\cite{Rusu_ICRA2011_PCL} and others released by their authors.

We measure root mean squared error (RMSE), mean absolute error (MAE), and the isotropic error (Error). Angular measurements are in units of degrees.

\subsection{Evaluation on ModelNet40}
\label{sec4.3}
\begin{figure*}[t]
    \centering
    \includegraphics[width=\linewidth]{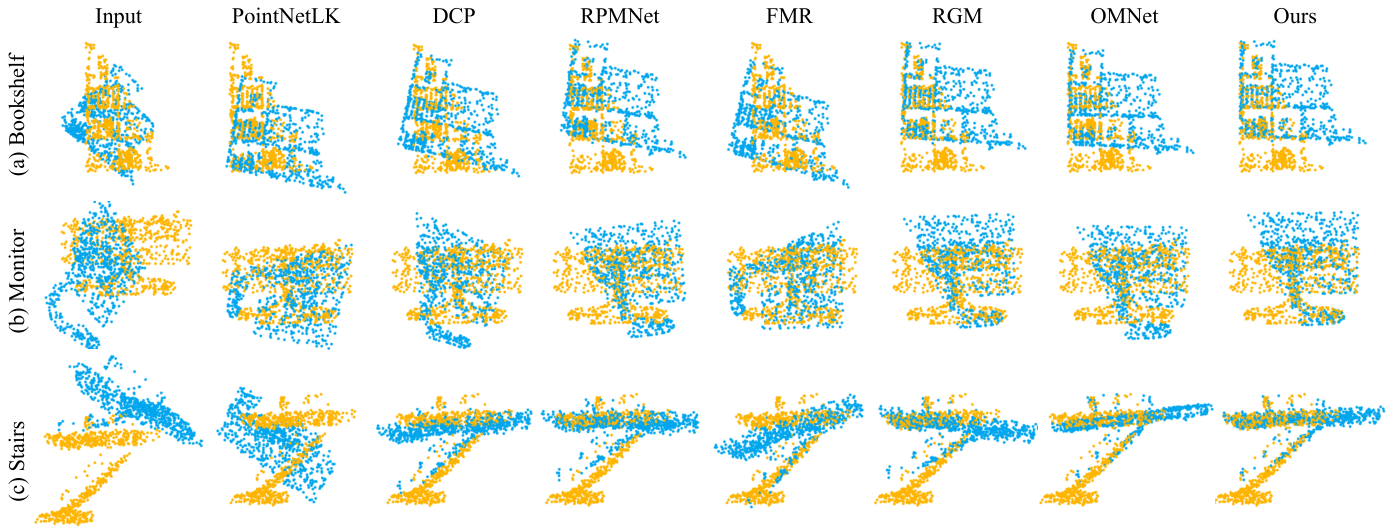}\\
    \caption{Qualitative examples on (a) Unseen objects, (b) Unseen categories, and (c) Gaussian noise.}
    \label{fig:qualitative_result}
\end{figure*}

\noindent\textbf{Unseen Objects. }
We first evaluate the models on the same categories. Note that the source and the reference point clouds of the \emph{TS} data have no exact correspondences in each other. Table~\ref{tab:main}(a) shows the results, where our method ranks first in all measures. We observe that Go-ICP and DeepGMR fail to obtain reasonable results on the partially visible data of RPMNet due to the partial manner, so that we do not report their results in Table~\ref{tab:main}. A qualitative comparison of the registration results can be found in Fig.~\ref{fig:qualitative_result}(a).

% \vspace{-0.3cm}
\noindent\textbf{Unseen Categories. }
To evaluate the generalization ability, we train the models on the first 14 categories and test on the test set of the remaining unseen categories. The results are summarized in Table~\ref{tab:main}(b). It can be found that all the learning-based methods consistently perform worse without training on the same categories. However, the traditional algorithms are not affected so much because that the handcrafted features are not sensitive to the shape variance. Our method outperforms its competitors in all metrics. A qualitative example of registration can be found in Fig.~\ref{fig:qualitative_result}(b).

\begin{table}[t]
    \normalsize
    \resizebox{\linewidth}{!}{%
    \begin{tabular}{rc|cccccccccccc}
        \toprule
            & &    PFI     &     GFI    &     SP     &     PFDL   &     TSL    & RMSE(\textbf{R}) & RMSE(\textbf{t}) & Error(\textbf{R}) & Error(\textbf{t}) \\
        \midrule
        & 1) &            &            &            &            &            & 4.921 & 0.0523 & 4.850 & 0.0562 \\
        & 2) & \Checkmark &            &            &            &            & 4.368 & 0.0459 & 3.982 & 0.0465 \\
        & 3) &            & \Checkmark &            &            &            & 4.792 & 0.0495 & 4.575 & 0.0532 \\
        & 4) & \Checkmark & \Checkmark &            &            &            & 4.427 & 0.0451 & 3.846 & 0.0448 \\
        & 5) & \Checkmark & \Checkmark & \Checkmark &            &            & 4.481 & 0.0469 & 3.981 & 0.0489 \\
        \multirow{-6}{*}{\rotatebox{90}{Single Branch}} & 6)  & \Checkmark & \Checkmark &  & \Checkmark & & 4.396 & 0.0427 & 3.804 & 0.0410 \\
        \midrule
        & 7) &            &            &            &            &            & 4.857 & 0.0507 & 4.624 & 0.0557 \\
        & 8)  & \Checkmark &            &            &            &            & 4.308 & 0.0435 & 3.799 & 0.0429 \\
        & 9) &            & \Checkmark &            &            &            & 4.220 & 0.0429 & 3.666 & 0.0419 \\
        & 10) & \Checkmark & \Checkmark &            &            &            & 4.023 & 0.0411 & 3.294 & 0.0387 \\
        & 11) & \Checkmark & \Checkmark & \Checkmark &            &            & 3.955 & 0.0401 & 3.183 & 0.0370 \\
        & 12) & \Checkmark & \Checkmark & \Checkmark & \Checkmark &            & 3.902 & 0.0399 & 3.082 & 0.0368 \\
        \multirow{-8}{*}{\rotatebox{90}{Dual Branches}} & 13) & \Checkmark & \Checkmark & \Checkmark & \Checkmark & \Checkmark & \textbf{3.841} & \textbf{0.0379} & \textbf{2.984} & \textbf{0.0336} \\
         \bottomrule
    \end{tabular}%
    }
    \caption{Ablation studies.}
    \label{tab:ablation_studies}
\end{table}

% \vspace{-0.3cm}
\noindent\textbf{Gaussian Noise. }
In this experiment, we evaluate the robustness to noise, which is always presented in real-world point clouds. We test on the unseen categories and add noise that independently sampled from $\mathcal{N}(0,0.01^{2})$ and clipped to $[-0.05,0.05]$ for each point. As shown in Table~\ref{tab:main}(c), all learning-based methods get worse with noises injected. Our method exhibits the best robustness compared to the others. An example result on noisy data is shown in Fig.~\ref{fig:qualitative_result}(c).

% \vspace{-0.3cm}
\noindent\textbf{Different Partial Manner. }
To valid the effectiveness to different partiality, we further test all the algorithms on the unseen categories used the partial manner in PRNet~\cite{wang2019prnet}. We retrain the leaning-based methods and the results are shown in Table~\ref{tab:different_partial_manner}. Since RPMNet and RGM use the different partially visible data from PRNet and IDAM, we evaluate them separately. Our method shows stronger robustness, and achieves the best performance.

\subsection{Evaluation on 7Scenes}
\label{sec4.4}
\begin{figure}[t]
    \centering
    \includegraphics[width=\linewidth]{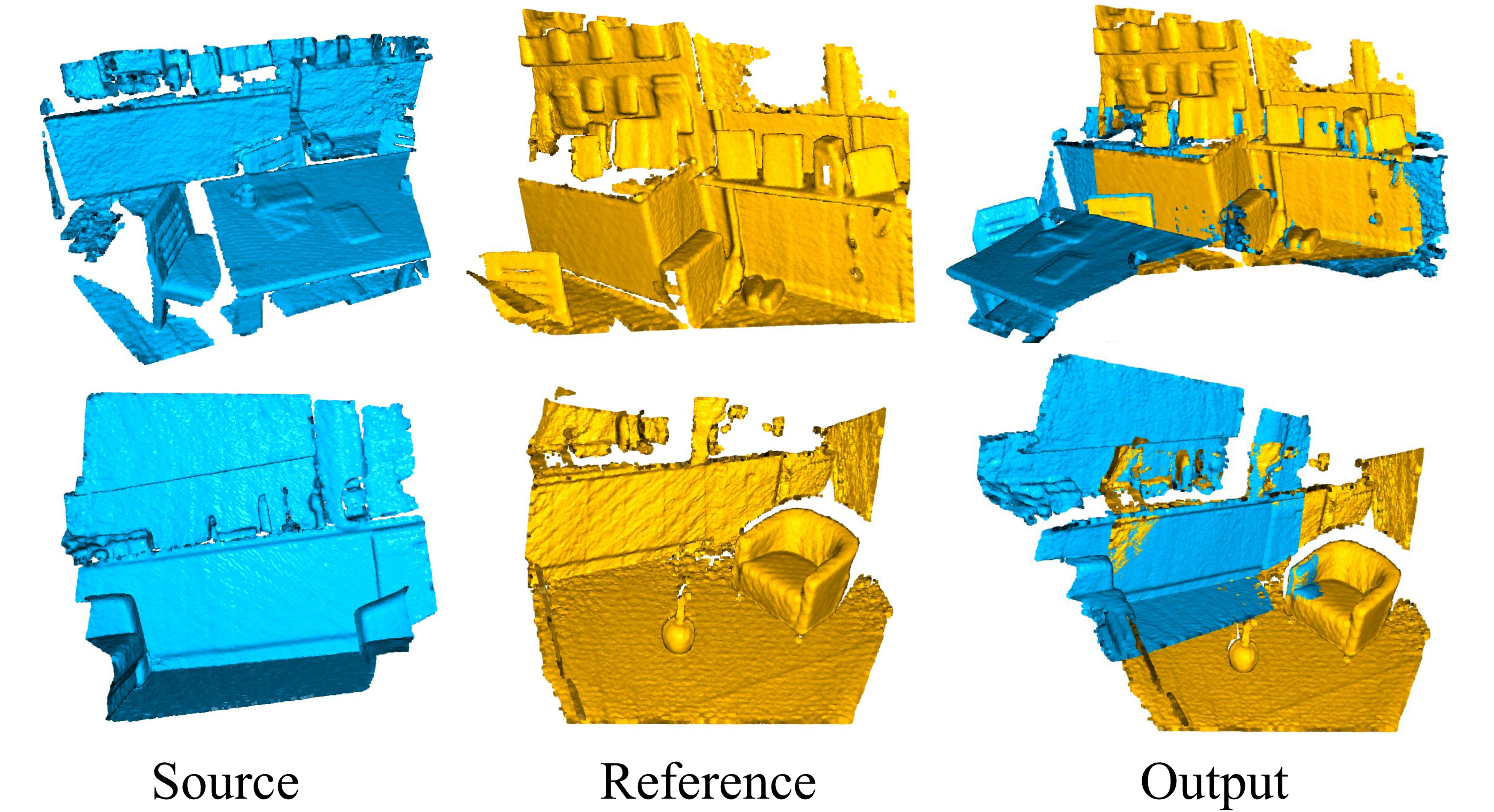}\\
    \caption{Qualitative results on 7Scenes (30$\sim$50\% overlap).}
    \label{fig:7scenes_qualitative_results}
\end{figure}

We conduct experiments on the 7Scenes dataset. All scans are randomly sampled to the point clouds with different density, results on which are shown in Fig.~\ref{fig:7scenes_diff_points}. D3Feat and PREDATOR show performance degradation as the point clouds become sparse. Our method exhibits robustness to the density variations. Although D3Feat and PREDATOR achieve higher precise with large inputs, it is worth to note that our method is at least 10 times faster than them. Fig.~\ref{fig:7scenes_qualitative_results} shows some qualitative results. Performance will degrade significantly with input points less than 1k.
% We further test our method with 250 points and obtain Error(\textbf{R})=13.307 and Error(\textbf{t})=0.0956, and performance will degrade significantly with input points less than that.
\begin{figure*}[t]
    \centering
    \includegraphics[width=\linewidth]{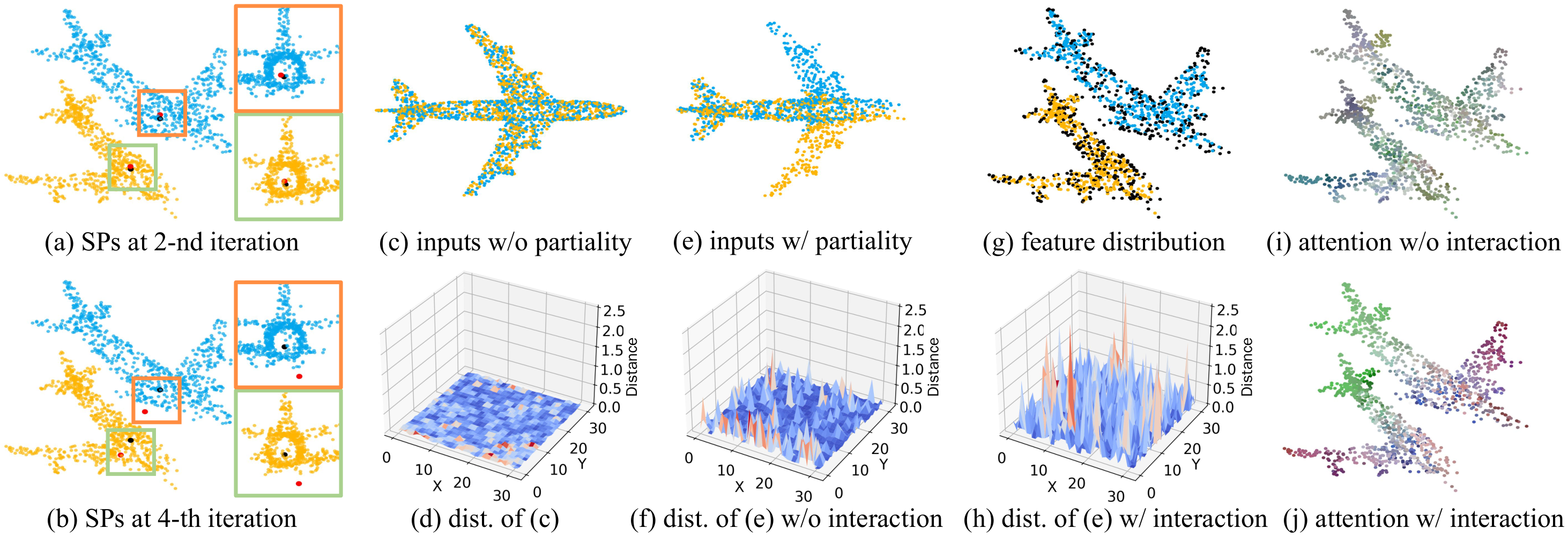}\\
    \caption{Visualization results in \S\ref{sec5.1} and \S\ref{sec5.2}. SP denotes saliency point and dist. means distance.}
    \label{fig:discussion}
\end{figure*}
\subsection{Ablation Studies}
\label{sec4.5}
% \vspace{-0.3cm}
\noindent\textbf{Dual Branches Structure. }
The dual branches structure (DB) is an important feature of our network. Here, we replace it with a single branch structure (SB), where two branches are shared weights. For fairness, we double the output channel of each convolution layer in SB to obtain a comparable number of parameters to DB. Comparing Row 1 with 7 in Table~\ref{tab:ablation_studies}, simply applying the DB structure only brings a slight improvement without additional supervising. However, only the DB structure can enable the ability of learning attentive features and enhance feature interactions.

% \vspace{-0.3cm}
\noindent\textbf{Multi-level Feature Interaction. }
The multi-level feature interaction is another important aspect. We compare the performances in the case of different combinations of the PFI and GFI. Comparing Row 1$\sim$4 in Table~\ref{tab:ablation_studies}, we can see that only the PFI improves the performance with a large margin. Since the same encoder is applied for the rotation and translation in SB, it may confuse the GFI with their information blended. However, comparing Row 7$\sim$10, both the PFI and GFI improve the performance in DB, and it achieves an extra improvement by combining them.

% \vspace{-0.3cm}
\noindent\textbf{Loss Functions. }
We verify the effectiveness of our Point-wise Feature Dropout Loss (PFDL) and Transformation Sensitivity Loss (TSL). Comparing Row 4 with 6, and Row 11 with 12 in Table~\ref{tab:ablation_studies}, PFDL improves the performance when applying on both SB and DB. Note that TSL can only be applied to DB. Comparing Row 12 with 13, it improves the performance with the TSL supervising, which supports the intuition that registration becomes more precise with the rotation and translation features more attentive to themselves, and it exists space difference between them.

% % \vspace{-0.3cm}
% \paragraph{Point-wise Feature Dropout Loss.}
% We also study the significance of our Point-wise Feature Dropout Loss (PFDL) by ablating it. Comparing Row 5 with 6, and Row 11 with 12 in Table~\ref{tab:ablation_studies}, the PFDL can marginally improves the performance on the test set. In addition, we notice that the network converges more rapidly with the PFDL supervising.

% \vspace{-0.3cm}
% \paragraph{Number of Iterations.}
% The model is trained with different iterations to show that how many iterations are needed. As illustrated in Figure~\ref{fig:number of iteration},  we can see that the most performance gains are in the first two iterations. We choose $N=4$ for the trade-off between the speed and effectiveness.

\section{Discussion}
\subsection{Translation Regression: Distance or SP?}
\label{sec5.1}
We first try to answer why estimating the saliency points (SP) is better. Intuitively, directly regressing the translation $\mathbf{t}$ implements two steps implicitly: (1) estimate the SP for the inputs; (2) compute the difference between them as $\mathbf{t}$, which can be described as $\mathbf{R}\mathbf{c}_{\scriptstyle \mathcal{X}}+\mathbf{t}=\mathbf{c}_{\scriptstyle \mathcal{Y}}$,
\begin{equation}
\small
\left\{ 
\begin{array}{ll}
\mathbf{R} \not\approx \mathbf{E} & \text{s.t. } \mathbf{R}\mathbf{c}_{\scriptstyle \mathcal{X}}=\mathbf{c}_{\scriptstyle \mathcal{X}},\; \mathbf{c}_{\scriptstyle \mathcal{X}} \leftrightarrow \mathbf{c}_{\scriptstyle \mathcal{Y}} \\
\mathbf{R} \approx \mathbf{E} & \text{s.t. } \mathbf{c}_{\scriptstyle \mathcal{X}} \leftrightarrow \mathbf{c}_{\scriptstyle \mathcal{Y}}
\end{array}\right.
\Rightarrow \,
\mathbf{t}=\mathbf{c}_{\scriptstyle \mathcal{Y}}-\mathbf{c}_{\scriptstyle \mathcal{X}}.
\end{equation}
Here, $\leftrightarrow$ denotes the corresponding relationship, and $\mathbf{E}$ means the identity matrix. It requires that SP are the rotation centers when $\mathbf{R} \not\approx \mathbf{E}$, otherwise $\mathbf{c}_{\scriptstyle \mathcal{X}}$ only needs to correspond to $\mathbf{c}_{\scriptstyle \mathcal{Y}}$. Considering it exists a large rotation gap between the inputs, the difference between their rotation centers is the translation distance, which is difficult to regress directly because the process of finding rotation centers is implicit. Comparing Row 4 with 5 in Table~\ref{tab:ablation_studies}, the hybrid features of the rotation and the translation interfere with the regression of SP in SB. However, comparing Row 10 with 11, regressing SP in DB improves the performance. Therefore, as an explicit goal, regressing SP helps to reduce learning difficulty. We visualize the rotation centers and the predicted SP in black and red respectively, as shown in Fig.~\ref{fig:discussion}(a-b). The predicted SP close to the rotation center when it exists a rotation gap between the inputs, otherwise merely form the correspondences.

\subsection{Partiality Perception Ability}
\label{sec5.2}
To explore the partiality perceptual ability of our method, we calculate the distances between the global features of the inputs. As shown in Fig.~\ref{fig:discussion}(f), it tries to extract the same features for the perturbed inputs without the feature interaction. In contrast to the intuition that concentrating on the overlapping regions, the network seems to learn how to utilize the partiality information to better register with the feature interaction, resulting in the larger distance between the global features, as shown in Fig.~\ref{fig:discussion}(h). 

Moreover, we visualize where the values in global feature come from when utilizing the feature interaction, and the source points are shown in black. As shown in Fig.~\ref{fig:discussion}(g), some of the non-overlapping points also have contributions because
the spatial position of a particular part relative
to the whole object can be used as prior information to help
with registration. Finally, we show the attentive areas of the global features with and without the feature interaction, where the colors denote different iterations, as shown in Fig.~\ref{fig:discussion}(i-j). With the data association, the network tends to focus on distinct areas at each iteration. In contrast, the areas are almost the same without the feature interaction, which reflects the lack of partiality perception.

\section{Conclusion} 
% \vspace{-1mm}
We have presented the FINet, an end-to-end global feature-based algorithm for adapting data association for the partial-to-partial point cloud registration. Our method possesses the multi-level feature interaction based on a dual branches structure, which enables efficient and effective early information exchange between the inputs. In addition, we design a transformation sensitivity loss and a point-wise feature dropout loss to learn attentive and distinctive features for the rotation and the translation respectively. Experimental results show the state-of-the-art performance and robustness of our method.

% \textbf{Acknowledgements. } This work is supported by the National Key R\&D Plan of the Ministry of Science and Technology (Project No.2020AAA0104400) and the National Natural Science Foundation of China (NSFC) under Grants No.62071097, No.61872067, No.62031009 and No.61720106004.

\section{Acknowledgments} 
This work was supported by the National Natural Science Foundation of China (NSFC) under Grants No.62071097, No.61872067, and No.61720106004.

\end{document}